\documentclass[letterpaper, 10 pt, conference]{ieeeconf}  
\usepackage{algorithm} 
\usepackage{algpseudocode} 
\usepackage{cite}

\IEEEoverridecommandlockouts                              

\usepackage{graphicx}
\usepackage{multirow}
\usepackage{booktabs}
\usepackage{amsmath}
\usepackage{arydshln}
\usepackage{tabularray}
\usepackage{array}
\usepackage{booktabs}
\usepackage{enumerate}
\usepackage{amssymb} 
\usepackage{hyperref}
\usepackage{xcolor}
\usepackage{threeparttable}
\usepackage{lscape}

\usepackage{makecell}
\usepackage[normalem]{ulem}

\usepackage{adjustbox}
\usepackage{tikz}
\usepackage{mathtools}

\title{\LARGE \bf
IMOST: Incremental Memory Mechanism with Online Self-Supervision for Continual Traversability Learning
}
\author{Kehui Ma, Zhen Sun, Chaoran Xiong, Qiumin Zhu, Kewei Wang, Ling Pei$^\dagger$ 
\thanks{Kehui Ma, Zhen Sun, Chaoran Xiong, Qiumin Zhu, Kewei Wang, Ling Pei are with School of Elcctronic Information and Electrical Engincering, Shanghai Jiao Tong Univcrsity, China. The work was supported in part by the National Nature Science Foundation of China (NSFC) under Grant No.62273229 and No.61873163 separately.}
\thanks{\textdagger\ indicates the corresponding authors: Ling Pei (ling.pei@sjtu.edu.cn)}%
}

\begin{document}

\maketitle
\thispagestyle{empty}
\pagestyle{empty}

\begin{abstract}

Traversability estimation is the foundation of path planning for a general navigation system. However, complex and dynamic environments pose challenges for the latest methods using self-supervised learning (SSL) technique. Firstly, existing SSL-based methods generate sparse annotations lacking detailed boundary information. Secondly, their strategies focus on hard samples for rapid adaptation, leading to forgetting and biased predictions. In this work, we propose IMOST, a continual traversability learning framework composed of two key modules: incremental dynamic memory (IDM) and self-supervised annotation (SSA). By mimicking human memory mechanisms, IDM allocates novel data samples to new clusters according to information expansion criterion. It also updates clusters based on diversity rule, ensuring a representative characterization of new scene. This mechanism enhances scene-aware knowledge diversity while maintaining a compact memory capacity. The SSA module, integrating FastSAM, utilizes point prompts to generate complete annotations in real time which reduces training complexity.  Furthermore, IMOST has been successfully deployed on the quadruped robot, with performance evaluated during the online learning process. Experimental results on both public and self-collected datasets demonstrate that our IMOST outperforms current state-of-the-art method, maintains robust recognition capabilities and adaptability across various scenarios. The code is available at \href{https://github.com/SJTU-MKH/OCLTrav.git}{https://github.com/SJTU-MKH/OCLTrav}.

\end{abstract}
\section{INTRODUCTION}


With the development of embodied AI technology, research on traversability estimation has increasingly focused on online continual learning for robots\cite{freyFastTraversabilityEstimation2023, mattamala01022024WildVisualNavigation2024, yoon01442044AdaptiveRobotTraversability2024}. Online continual learning involves the process of continuous learning and adaptation from real-time data streams on robotic platforms, which are constrained by limited computational resources and storage capacity. Currently, geometry-based approaches assess traversability by analyzing slope, flatness, and height information, which allows for the establishment of consistent rules in structured environments\cite{caoAutonomousExplorationDevelopment2022}. In contrast, unstructured environments often feature irregular and dynamic  obstacles and terrain. This makes it challenging to establish consistent and reliable traversability rules. Consequently, semantic-based approaches have attracted considerable interest, as sophisticated semantic features allow for precise traversability estimation in unstructured environments\cite{seo01112024LearningOffRoadTerrain2023}. Nevertheless, semantic-based supervised learning approaches rely heavily on labeled data\cite{freyLocomotionPolicyGuided2022a}. Furthermore, obtaining high-quality, dense annotations in diverse environments is expensive and unscalable. To address this issue, self-supervised learning techniques have been developed. Our work builds upon and extends these methods.

\begin{figure}[t]
    \centering
    \includegraphics[width=0.45\textwidth]{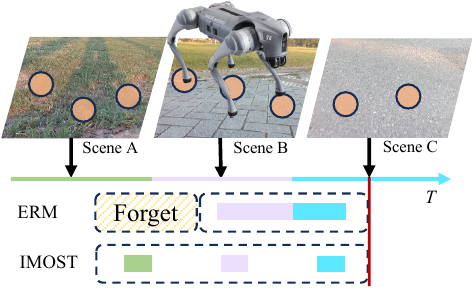}
    \caption{\textbf{Continual traversability learning}. Under limited resource, previous experience replace methods (ERM) did not take into account the semantic distribution across multiple scenarios, leading to imbalanced data and catastrophic forgetting.Ours IMOST incrementally and evenly preserves data related to different scenes. }
    \label{fig:problem}
\end{figure}

SSL-based methods project motion trajectories onto images to obtain sparse traversability labels\cite{zurnSelfSupervisedVisualTerrain2021, castroHowDoesIt2023, dongshinkimTraversabilityClassificationUsing2006}. While incomplete annotation information increases the difficulty of model learning, previous research\cite{wangContinualLearningRetrieval2022} uses clustering techniques to derive denser labels, but the boundaries of between traversable and non-traversable regions remain ambiguous. Recently, the advanced SAM tool has been introduced, utilizing point prompts to acquire detailed annotations \cite{jungVSTRONGVisualSelfSupervised2024}. However, due to the high computational demands of the SAM model, its real-time performance is constrained when deployed on resource-limited mobile platforms. Hence, existing methods struggle with achieving both annotation completeness and real-time processing.

Meanwhile, domain distribution shift and imbalanced data remain persistent issue in continual learning, as discussed in \cite{mccloskeyCatastrophicInterferenceConnectionist1989}. Neural network adapting to new environment tends to forget previously learned knowledge. WVN \cite{freyFastTraversabilityEstimation2023} proposes an online self-supervised approach that can adapt to localized scenes based on human demonstrations. However, the model's performance still suffers from degradation across diverse environment. Hyung-Suk Yoon et al.\cite{yoon01442044AdaptiveRobotTraversability2024} builds on experience replay with uncertainty, considering the uncertainty of the dataset to improve adaptation. Yet, this approach does not specifically focus on achieving data balance across various scenes.

To address these limitations, we propose IMOST, a novel approach designed to enhance continual traversability learning. IMOST comprises three key modules: incremental dynamic memory (IDM), self-supervised annotation (SSA), and inference network. The human brain has the ability to recognize and store new knowledge. Through knowledge consolidation, the brain retains past knowledge while also learning new information\cite{squireMemoryConsolidation2015}. Inspired by this mechanism, IDM increases new cluster as the scenes evolve according to information expansion criterion, which helps in identifying new scenes and addressing the issue of domain distribution shift. And capacity limitations are applied for clusters to ensure data balance across different domain distributions. Self-supervised annotation assisted by FastSAM achieves complete annotations in real time. FastSAM applies a new architecture that leverages the computational efficiency of CNNs, achieving a comparable performance with the SAM method at 50× higher run-time speed\cite{zhaoFastSegmentAnything2023}. In this paper, our main contributions can be summarised as follows:
\begin{enumerate}
    \item We present an efficient and real-time self-supervised annotation pipeline that integrates proprioceptive information and FastSAM, without requiring extensive computational resources. 

    \item We propose a scene-aware incremental dynamic memory inspired by human memory mechanisms and design \textbf{information expansion criterion} to  identify new domain distribution.
    \item Real-world experiments demonstrate that our method outperforms the baseline and significantly improves the ability to continually learn traversability across diverse scenes. And we will open source the code to promote further advancement in the field.
\end{enumerate}

\section{RELATED WORK}

\subsection{self-supervised Learning for Traversability}
Self-supervised labeling techniques are continuously improving and evolving to streamline the data annotation workload. Current methods utilize various sensors to assist in the generation of self-supervised data and produce traversability scores\cite{seoScaTEScalableFramework2023, fan*STEPStochasticTraversability2021, rebuffiICaRLIncrementalClassifier2017,CORL23-karnan}. The underlying method remains consistent: odometry data is mapped onto images\cite{wellhausenWhereShouldWalk2019, dahlkampSelfsupervisedMonocularRoad2006, caiRiskAwareOffRoadNavigation2022a}. In the work of \cite{freyFastTraversabilityEstimation2023}, Supervision and Mission Graphs were constructed to map local footprints onto images, resulting in sparse supervision signals. At the raw data level, the clustering method SLIC\cite{achanta00292012SLICSuperpixelsCompared2012} was used to obtain denser labels, and the confidence was derived from the distribution of traversability data. Matias Mattamala et al.\cite{mattamala01022024WildVisualNavigation2024} improved upon this by clustering at the feature level, yielding more reliable results. However, the boundary between traversable and non-traversable regions remains ambiguous. Recently, Vision Foundation Models\cite{kirillovSegmentAnything2023a, oquabDINOv2LearningRobust2023, caron01252021EmergingPropertiesSelfSupervised2021, hamilton2022unsupervised} have garnered significant attention. Sanghun Jung et al.\cite{jungVSTRONGVisualSelfSupervised2024} noted the potential of Vision Foundation Models for annotation tasks and applied SAM to wild environments to obtain positive and negative sample boundaries. Yunho Kim et al.\cite{kimLearningSemanticTraversability2024} integrated SAM for urban environments. However, due to SAM’s high computational demands, the above methods cannot be deployed on robots in real time, requiring pre-collected data to train the model. Our approach utilizes FastSAM\cite{zhaoFastSegmentAnything2023} to enable real-time deployment. And this work primary focuses on the continuous learning of traversability tasks.

\subsection{Continual Learning for Traversability}
Embodied robots are expected to meet higher standards and intelligent system with continuous learning capabilities can better adapt to the dynamic changes in the world. Continuous learning involves the ability to update, generalize, and accumulate knowledge similarly to human\cite{liuAligningCyberSpace2024a}. However, alternating between old and new tasks often leads to catastrophic forgetting, where learning new tasks impairs performance on previously learned ones\cite{wangComprehensiveSurveyForgetting2023}. In traversability learning, different scenes present varying feature distributions but share a common binary label indicating whether a point is traversable. This scenario exemplifies a typical Domain-Incremental Learning (DIL) problem \cite{wangComprehensiveSurveyContinual2024}. 

It’s important to note that online data streams do not provide unique identifiers for each scene. WVN\cite{freyFastTraversabilityEstimation2023} either rely on infinite storage or retain only the most recent data for random sampling, neglecting the informativeness of different samples, which does not effectively address the forgetting problem. Hyung-Suk Yoon et al.\cite{yoon01442044AdaptiveRobotTraversability2024} highlighted domain shift issues and employed heuristic selection to focus on hard sample detection based on experience replay mechanisms. While this approach improves adaptability to local scenes, it does not fully resolve the forgetting problem due to domain shifts in memory caused by excessive replacement of well-performing scene data. In other research domains, some approaches such as Mean-of-Feature \cite{davariProbingRepresentationForgetting2022} allocate an equal number of samples to each class center, while iCaRL \cite{zhuLearningbasedTraversabilityCostmap2024} selects samples closest to each cluster center. Fei Ye et al.\cite{yeTaskFreeContinualGeneration2024} increases the number of clusters based on the differences in latent vectors for the task. A more complex approach, DRI\cite{wangContinualLearningRetrieval2022}, additionally trains a generative model to recover old samples. Considering both computational complexity and effectiveness, we adopt a memory-based feature replay mechanism designed to store keyframe features for each scene, ensuring a balanced memory distribution across multiple scenes.

\begin{figure*}[ht]  
    \centering
    \includegraphics[width=\textwidth]{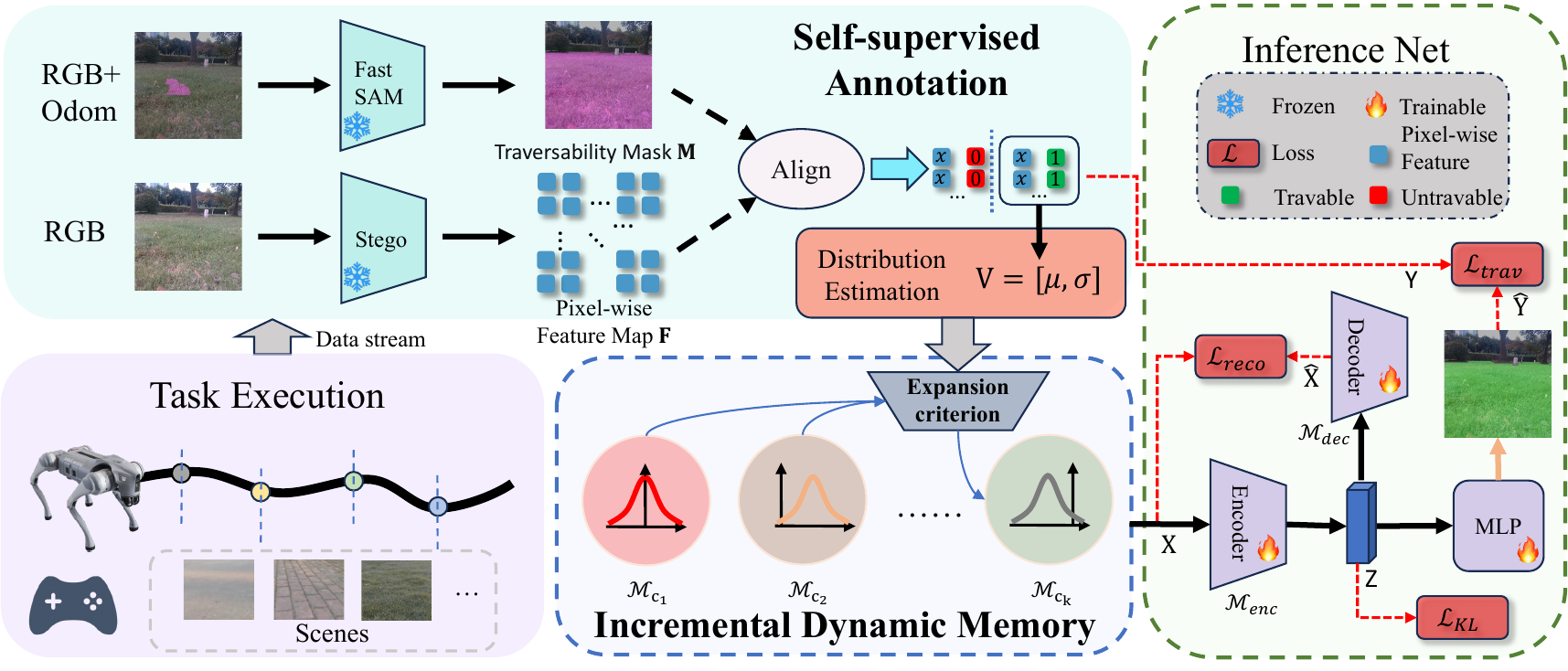}
    \caption{\textbf{The overview of IMOST}. \textbf{FastSAM} generates the traversability mask, with the traversable regions highlighted in purple in the image. Stego produces the feature map. The mask and feature map are combined to compute the feature distribution of the traversable areas. Then \textbf{IDM} updates the cluster according to \textbf{expansion criterion} and heuristically selects data to be input into the inference network for training.}
    \label{fig:structure}
\end{figure*}
\section{METHOD}

\subsection{Problem Definition and System Overview }

\textit{Problem Definition}: In the context of continual learning for traversability, we aim to achieve accurate and real-time predictions, while ensuring the model can continuously learn and adapt to new environments without performance degradation. Suppose we have a dataset $\mathcal{D}$ consisting of $\mathcal{N}$ sequences, i.e.  $\mathcal{D}$ =$\{ \mathcal{D}_1, \mathcal{D}_2, \dots, \mathcal{D}_\mathcal{N}\}$.  Each $\mathcal{D}_i$ represents a collection of data pairs $(\textbf{x}_j^i, y_j^i)$ from the $i$-th sequence, where $\textbf{x}_j^i$ and $y_j^i$ respectively denote the pixel-wise feature and the corresponding traversability label. The $\textbf{x}_j^i$ is from feature map $\textbf{F}$ generated by semantic network. Traversability label is part of traversability Mask $\textbf{M}$. Within these $\mathcal{N}$ sequences, there are $\mathcal{C}$ distinct scenes, with $\mathcal{C} \leq \mathcal{N}$, indicating that some scenes may be repeated across the sequences. 

\textit{System overview}: The proposed method is illustrated in Figure \ref{fig:structure}. This pipeline consists of four modules: task execution, self-supervised annotation, incremental dynamic memory, and inference network.
\textit{Task execution} involves following human control commands to navigate through traversable scenes for online learning. \textit{Self-supervised annotation} (Sec. \ref{self-supervised annotation}) includes two pre-trained networks, FastSAM and Stego, which generate traversability mask and pixel-wise feature map. \textit{Incremental dynamic memory} (Sec. \ref{IDM}) automatically adds new clusters and updates memory data as the system operates online. Finally, \textit{inference network} (Sec. \ref{Inference Network}) employs an autoencoder architecture, trained with memory data to learn and predict traversability.

\subsection{Self-supervised Annotation}
\label{self-supervised annotation}

SSA eliminates the need for manual labeling. During the human demonstration process, the robot’s trajectory is used to identify traversable areas. Sparse odometry is transformed into the image according to the results of robot calibration, generating the self-supervised signal. Then this signal is served as point prompts of FastSAM to infer comprehensive traversability masks.

\textbf{Robot calibration.} The robot is equipped with two primary sensors: a camera and a lidar. Before online data collection, the calibration among the robot, camera, and lidar needs to be completed. Using the existing calibration tool, the transformation $\mathbf{T}_{lidar}^{cam}$ between the camera and lidar can be computed. Additionally, the transformation $\mathbf{T}_{base}^{fc}$ between the robot's foot center and base is obtained through manual measurements, as well as the transformation $\mathbf{T}_{base}^{cam}$ between the camera and base. Subsequently, $\mathbf{T}_{fc}^{cam}$ and $\mathbf{T}_{lidar}^{base}$ can be derived. 

\textbf{Generating self-supervised data.} For each image, odometry within  $d_{max}$ of this image points are used as valid self-supervised points. We obtain the robot’s odometry using FAST-LIO \cite{xuFASTLIOFastRobust2021}, which provides the position and orientation information $\mathbf{T}_{lidar}^{odom}$. Then, valid self-supervised points $\textbf{P}_{odom}$ can be projected on the corresponding image to generate sparse annotations $\textbf{P}_{proj}$:
\begin{equation}
\mathbf{P}_{proj} = \mathbf{T}^{cam}_{fc}\mathbf{T}^{fc}_{base}\mathbf{T}^{base}_{lidar}\mathbf{T}^{lidar}_{odom}\mathbf{P}_{odom},
\label{eq:footprint_projection}
\end{equation}

Given the sparse annotations $\textbf{P}_{proj}$, point prompts of FastSAM are used to generate the traversability mask $\textbf{M}$ comparable to those created by human annotators. FastSAM utilizes a lightweight architecture optimized for rapid processing, making it ideal for real-time applications. 

Inspired by \cite{mattamala01022024WildVisualNavigation2024}, we use the pre-trained model stego\cite{hamiltonUnsupervisedSemanticSegmentation2022} as the backbone for feature extraction. This model extracts a pixel-wise feature map \textbf{F} with the same size as the image \textbf{I}. The feature map \textbf{F} and traversability mask \textbf{F} are then matched to generate supervision data pairs $(\textbf{x}_j, y_j)$.

\subsection{Incremental Dynamic Memory}
\label{IDM}
To adapt to diverse scenes, we aim to establish a scene-aware balanced memory buffer. We propose the \textbf{information expansion criterion} to identify new scenes and subsequently add a new cluster. The sample selection policy is adjusted to fit this mechanism.

\textbf{Memory expansion and updating.} For each image, as detailed in \ref{self-supervised annotation}, the pixel-wise feature map $\textbf{F}$ and corresponding labels $\textbf{M}$ are obtained. The image node $\mathcal{N}(\textbf{F}, \textbf{M}, \textbf{V})$ is stored in memory, where $\textbf{V}$ is the vector of [$\mathbf{\mu(F[M])}$, $\mathbf{\sigma(F[M])}$]. Here, $\mu$ and $\sigma$ represent the mean and standard deviation functions, respectively. The vector $\textbf{V}$ characterizes the feature distribution of the traversable region.

\begin{figure}[t]
    \centering
    \includegraphics[width=0.45\textwidth]{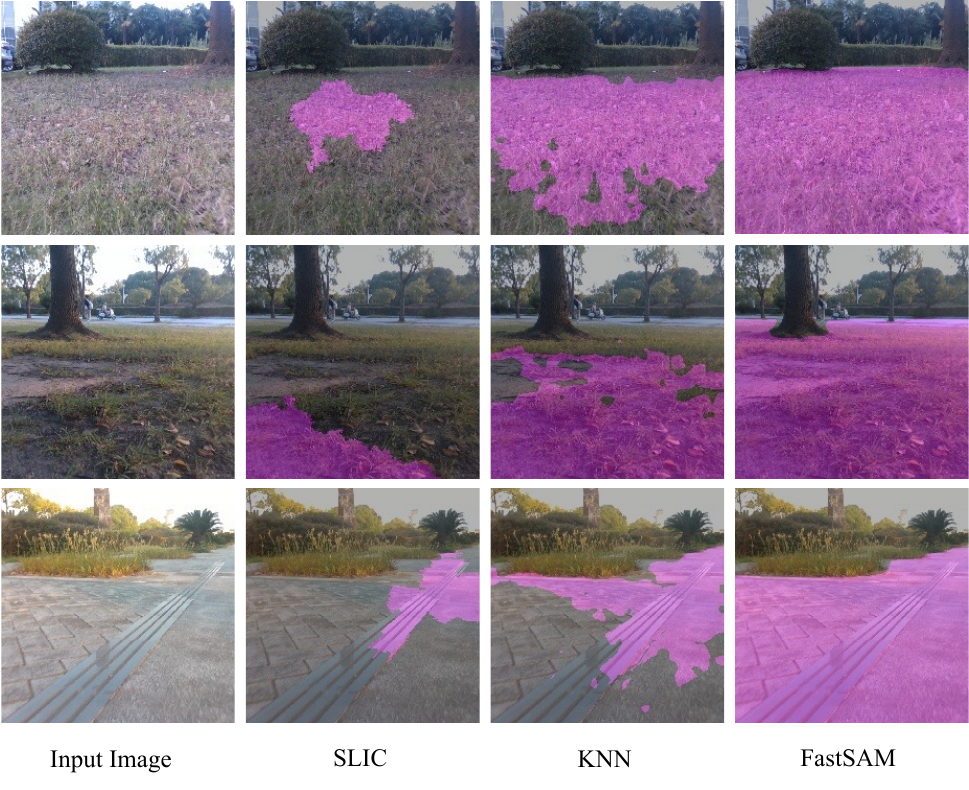}
    \caption{Comparison performance of self-supervised methods for 3 example images. WVN expands sparse labels by associating similar semantics.But the boundary segmentation between traversable and non-traversable areas is still not clear enough. Our SSA achieves nearly complete traversability semantic annotation.}
    \label{fig:annotation}
\end{figure}

Inspired by \cite{yeTaskFreeContinualGeneration2024}, we use the Jensen–Shannon divergence as the distance function between feature distribution vectors:
\begin{equation}
D_{JS}(V_1||V_2) = \frac{1}{2} KL(V1||V_2) + \frac{1}{2} KL(V_2||V_1),
\label{eq:DJS}
\end{equation}
where KL is Kullback–Leibler divergence. And it can be seen that $D_{JS}$ is a symmetrical statistical measure.

Our memory $\mathcal{M}_c$ consists of multiple clusters, denoted as \(\{\mathcal{M}_{c_1}, \mathcal{M}_{c_2}, \ldots, \mathcal{M}_{c_k}\}\). Each cluster $\mathcal{M}_{c_i}$ can store up to \(n_{max}\) image nodes and is characterized by the vector $V_{c_i}=[\mu(\mathcal{M}_{c_i}[\textbf{F}]), \sigma(\mathcal{M}_{c_i}[\textbf{F}])]$, which indicates the mean and standard deviation of the traversability features for all nodes in the $i$-th cluster. Then, the \textbf{information expansion criterion} for memory expansion is defined as follows:
\begin{equation}
min\{D_{JS}(V_{new}, V_{c_1}),...,D_{JS}(V_{new}, V_{c_k})\} \geq \lambda,
\label{eq:condition}
\end{equation}
where $V_{new}$ represents the feature distribution of new image node. If the minimum distance between this distribution and those of the existing clusters exceeds the threshold $\lambda$, a new cluster will be created. Otherwise, the nearest cluster incorporates the new image node and updates its representation.  The update strategy retains both the most similar and most diverse image nodes to the cluster center, ensuring a diverse representation of features within the cluster.

\textbf{Sampling policy.} Unlike existing methods such as uniform sampling and uncertainty-aware sampling \cite{yoon01442044AdaptiveRobotTraversability2024}, our sampling strategy incorporates both balanced sampling across clusters and uncertainty-aware sampling. Imbalanced data across different semantic categories during continual learning may lead to biased prediction. Hence, we assign an equal sampling probability of $\frac{1}{k}$ to each of the k clusters. Within the cluster, the weighting coefficient is inversely proportional to the number of nodes, ensuring that clusters with fewer nodes receive higher sampling weights. 

Additionally, we use reconstruction loss as weights of uncertainty. Within each cluster, all losses are normalized, and weights are assigned to each node accordingly, which is a low-complexity operation. The large loss indicates that the network’s memory performance for the node is poor, thus requiring more frequent sampling and training.

\subsection{Anomaly Learning}
\label{Inference Network}


Our inference network is composed of an variational autoencoder (VAE) and a multilayer perceptron (MLP). The encoder $\mathcal{M}_{enc}$ compresses the input data into a latent space which is then used by the MLP to predict traversability. The decoder $\mathcal{M}_{dec}$ is designed to reconstruct the original data from the latent space. And we design three loss functions to enhance the performance of the model.

\textbf{Reconstruction loss.} The key insight is that the model only learns  traversability data, so data points that do not conform to traversability distribution are considered anomalies. It's achieved by minimizing the mean squared error:
\begin{equation}
\mathcal{L}_{reco}(\textbf{X}, \hat{\textbf{X}}) = \frac{1}{N} \sum_{i=1}^{N}(\textbf{X}_i - \hat{\textbf{X}}_i)^2,
\label{eq:loss1}
\end{equation}
where $\textbf{X}$ is pixel-wise feature of traversability and $\hat{\textbf{X}}$ is recontruction by VAE.

\textbf{Traversability loss.} This objective is to fit the self-supervised labels:
\begin{equation}
\mathcal{L}_{trav}(\textbf{Y}, \hat{\textbf{Y}}) = BCELoss(\textbf{Y}, \hat{\textbf{Y}}), 
\label{eq:loss2}
\end{equation}
where, \textbf{Y} is self-supervised labels and $\hat{\textbf{Y}}$ is predicted by $MLP(\mathcal{M}_{enc}(\textbf{X})))$.

\textbf{Regularization loss.} The purpose of this loss is to minimize the discrepancy between the prior and posterior distributions to ensure that the model makes minimal changes:
\begin{equation}
\mathcal{L}_{KL}(\textbf{Z}_{k}, \hat{\textbf{Z}}_{k}) = KL(\textbf{Z}_k||\hat{\textbf{Z}}_k), 
\label{eq:loss3}
\end{equation}
where $\textbf{Z}_k$ is the latent vector, and $\hat{\textbf{Z}}_k$ is generated by 
$\mathcal{\mathcal{M}}_{enc}(\mathcal{M}_{dec}(\textbf{Z}_k))$

The final combined loss function is therefore defined as follows:
\begin{equation}
\begin{multlined}
    \mathcal{L}_{total} = \omega_1 \mathcal{L}_{reco}(\textbf{X}, \hat{\textbf{X}}) + \omega_2 \mathcal{L}_{trav}(\textbf{Y}, \hat{\textbf{Y}}) \\ + \omega_3 \mathcal{L}_{KL}(\textbf{Z}_{k}, \hat{\textbf{Z}}_{k}),
\end{multlined}
\end{equation}

\section{EXPERIMENTAL RESULTS}

\subsection{Platform Description}
\label{expA}
The IMOST is deployed on the Unitree Go2 legged robot. This robot is equipped with the high-performance computation platform-Orin AGX. The main perception sensors include a Livox Mid-360 lidar and a D435i depth camera. By running FAST-LIO (Fast LiDAR-Inertial Odometry), the robot could accurately estimate its position and orientation. The D435i depth camera captures color images.

\subsection{Dataset Description}
\label{expB}
To evaluate our proposed method, we employ two datasets: a public dataset provided by \cite{freyFastTraversabilityEstimation2023} and a self-collected dataset covering various terrain. The public dataset includes three distinct sequences from different scenarios, while the self-collected dataset captures multiple scenarios within a school environment. In Table \ref{table:dataset}, the detailed information of the datasets is analyzed. 

\textbf{Public dataset}: Sequence1 is collected from an indoor environment, containing three traversability semantics: floor, white carpet, and green carpet. Sequence2 primarily features scenes from Bahnhofstrasse, Zürich’s main downtown street. Sequence3 comprises a mix of paved roads and yard scenes.

\begin{figure}[t]
    \centering
    \includegraphics[width=0.47\textwidth]{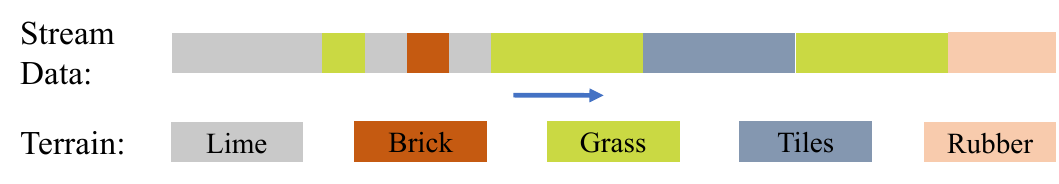}
    \caption{Detailed data flow. There are multiple scenes, and the data is imbalanced across different scenes.}
    \label{fig:datastream}
\end{figure}

\textbf{Self-collected dataset}: This dataset includes diverse terrains such as grasslands and various types of roads, including lime roads, brick roads, tiled roads, and others. The detailed scene information is illustrated in Fig. \ref{fig:datastream}. We evaluate our approach and perform ablation experiments on this dataset.

\begin{table}[t]
\centering
\caption{Dataset Overview}
\begin{tabular}{c|c|c|c|c} 
\specialrule{0.3pt}{0pt}{0pt} 
\toprule
Dataset    & Duration  & Distance & \#Train set& \#Test set \\ 
\midrule
\midrule
WVN-Seq1\cite{freyFastTraversabilityEstimation2023}        & 134s     &  160m    & 154   & 50           \\
WVN-Seq2\cite{freyFastTraversabilityEstimation2023}         & 106s     &  320m    & 212      & 50     \\
WVN-Seq3\cite{freyFastTraversabilityEstimation2023}         & 100s     &  100m    & 120      & 50         \\
Self-collected        & 740s    &  1108m    & 1168 & 694 \\
\bottomrule
\specialrule{0.3pt}{0pt}{0pt} 
\end{tabular}
\label{table:dataset}
\end{table}

\begin{table}[b]
\centering
\caption{Comparision of Self-Supervised Annotations}
\begin{threeparttable}{
\begin{tabular}{c|c|c|c|c} 
\specialrule{0.3pt}{0pt}{0pt} 
\toprule
Method      & params  & FPS & Time[ms] & IoU  \\ 
\midrule
\midrule
SLIC        & -     &  \textbf{125}  & 8.0  & 0.1657           \\
KNN         & -     &  45     &  22.2  & 0.6468            \\
FastSAM     & 145M     &  99  &  10.1  & \textbf{0.9927}  \\
\bottomrule
\specialrule{0.3pt}{0pt}{0pt} 
\end{tabular}
\begin{tablenotes}
\footnotesize
\item[] "-" denotes that non-deep learning methods do not include model parameters.
\end{tablenotes}
}
\end{threeparttable}
\label{table:study1}
\end{table}

\subsection{Compared Methods and Evaluation Metrics}
\label{expC}
This paper primarily focuses on the continual learning of traversablity across multiple scenarios. The latest online stream learning approach, WVN\cite{freyFastTraversabilityEstimation2023}, serves as the baseline. We also configure several variants of the IMOST to demonstrate the effectiveness of our modules:

\begin{itemize}[]
    \item \textbf{WVN}: This online stream learning method has no storage constraints and relies on the self-supervised labels generated from footprint mapping and KNN clustering.
    \item \textbf{IMOST w.o. IDM}: In this variant of our IMOST method, IDM is replaced by limited memory. The storage module operates as a first-in, first-out (FIFO) queue, with a fixed size limit of 150.
    \item \textbf{IMOST w.o. SAM}: In this version of IMOST, the annotation data uses segment-wise features derived from KNN cluster centers instead of SAM.
    
    \item \textbf{IMOST}: Our full method, trained with SAM self-supervised annotations and incremental dynamic memory (IDM). We set the threshold $\lambda$ to 1, with each cluster’s capacity limited to 20.
    
\end{itemize}


\begin{table}[t]
\centering
\caption{Evaluation Results on Public Dataset}
\begin{threeparttable}{
\resizebox{\columnwidth}{!}{
\begin{tabular}{c|c|c|c|c|c|c} 
\specialrule{0.3pt}{0pt}{0pt} 
\toprule
Dataset            & Method    & AUROC$\uparrow$ & $\beta$$\downarrow$   & F1$\uparrow$    & IoU$\uparrow$  & \#Cluster   \\ 
\midrule
\midrule
WVN-Seq1               & WVN       & 0.9766  & 0.3567    & 0.8844& 0.7997& *           \\
\multicolumn{1}{l|}{}   & IMOST  & \textbf{0.9979}  & \textbf{0.0577}    & \textbf{0.9774}& \textbf{0.9529}& 3           \\
\midrule
WVN-Seq2               & WVN       & 0.9977  & 0.3925    & 0.9621& 0.9282& *           \\
\multicolumn{1}{l|}{}   & IMOST  & \textbf{0.9989}  & \textbf{0.0778}    & \textbf{0.9830}& \textbf{0.9666}& 1           \\
\midrule
WVN-Seq3               & WVN       & 0.9943  & 0.4546    & 0.8905& 0.8052& *           \\
\multicolumn{1}{l|}{}   & IMOST  & \textbf{0.9963}  & \textbf{0.0405}    & \textbf{0.9739}& \textbf{0.9485}& 2           \\
\bottomrule
\specialrule{0.3pt}{0pt}{0pt} 
\end{tabular}
}
\begin{tablenotes}
\footnotesize
\item[] "*" indicates that the WVN model does not include the this metric.
\end{tablenotes}
}
\end{threeparttable}
\label{table:study2_metric}
\end{table}

Similar to the approach\cite{seo01112024LearningOffRoadTerrain2023}, we evaluate the traversability estimation as a binary classification problem. The model's performance is assessed using several metrics, including AUROC, precision, recall, F1-score, and IoU. AUROC provides a robust measure of model performance, particularly in imbalanced class scenarios, as it evaluates the model across all possible threshold values. The optimal decision threshold $\alpha$ is determined by finding the point on the ROC curve closest to (0,1), using the Euclidean distance $Norm(FPR, 1-TPR)$. We evaluated the bias $\beta$ of the optimal decision threshold using the following function:
\begin{equation}
\beta = |0.5 - \alpha|
\label{eq:bias}
\end{equation}

\begin{figure}[t]
    \centering
    \includegraphics[width=0.4\textwidth]{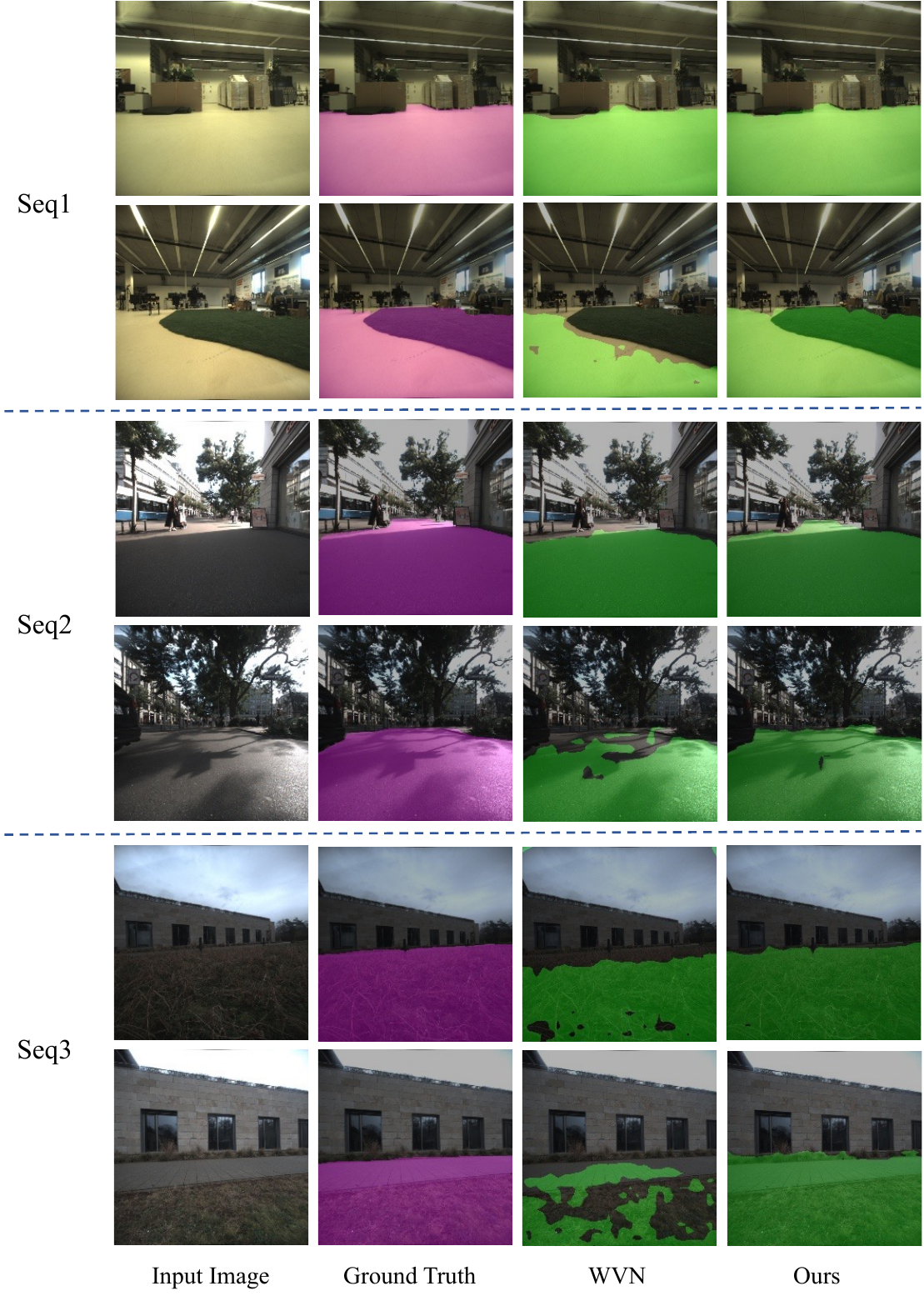}
    \caption{Qualitative results on public dataset.}
    \label{fig:study2_demo}
\end{figure}

\subsection{Experimental Result}
\label{expD}
\subsubsection{Study 1} Evaluation of Self-Supervised Annotation. This study compares our approach to WVN\cite{freyFastTraversabilityEstimation2023, mattamala01022024WildVisualNavigation2024} in terms of real-time performance and annotation completeness. The SLIC\cite{achanta00292012SLICSuperpixelsCompared2012} is a superpixel segmentation algorithm designed to partition an image into multiple regions (superpixels) that are homogeneous in color and spatial location. Superpixels that intersect with the footprint are classified as traversable. KNN is used for feature clustering with 20 cluster centers and clusters intersecting with the footprint are classified as traversable. Our approach, however, directly utilizes footprint points to generate the traversable mask via FastSAM’s point prompts. Comparison results on the AGX computing platform, shown in Table \ref{table:study1}, are based on testing with 100 images from the public dataset sequences. As shown, FastSAM outperforms the other methods in annotation accuracy and achieves a frame rate of 99 FPS.

\begin{table}[t]
\centering
\caption{Evaluation and ablation Results on Self-Collected Dataset}
\resizebox{\columnwidth}{!}{
\begin{tabular}{c|c|c|c|c|c} 
\specialrule{0.3pt}{0pt}{0pt} 
\toprule
 Method    & AUROC$\uparrow$      & F1$\uparrow$       & Precision$\uparrow$  & Recall$\uparrow$    & IoU$\uparrow$  \\ 
\midrule
\midrule
WVN        & 0.7564 & 0.6812     & \textbf{0.9961}    & 0.5176  & 0.5221          \\
IMOST        & \textbf{0.9725} & \textbf{0.9725}   & 0.9878     & \textbf{0.9578}   &  \textbf{0.9463}     \\
\midrule
IMOST w.o. IDM  & 0.9589  & 0.9319   & 0.9888     & 0.8812    & 0.8775             \\
IMOST w.o. SAM  & 0.9003  & 0.7548   & 0.9949     & 0.6081    & 0.6073              \\
\bottomrule
\specialrule{0.3pt}{0pt}{0pt} 
\end{tabular}
}
\label{table:study3_metric}
\end{table}

\begin{figure}[t]
    \centering
    \includegraphics[width=0.45\textwidth]{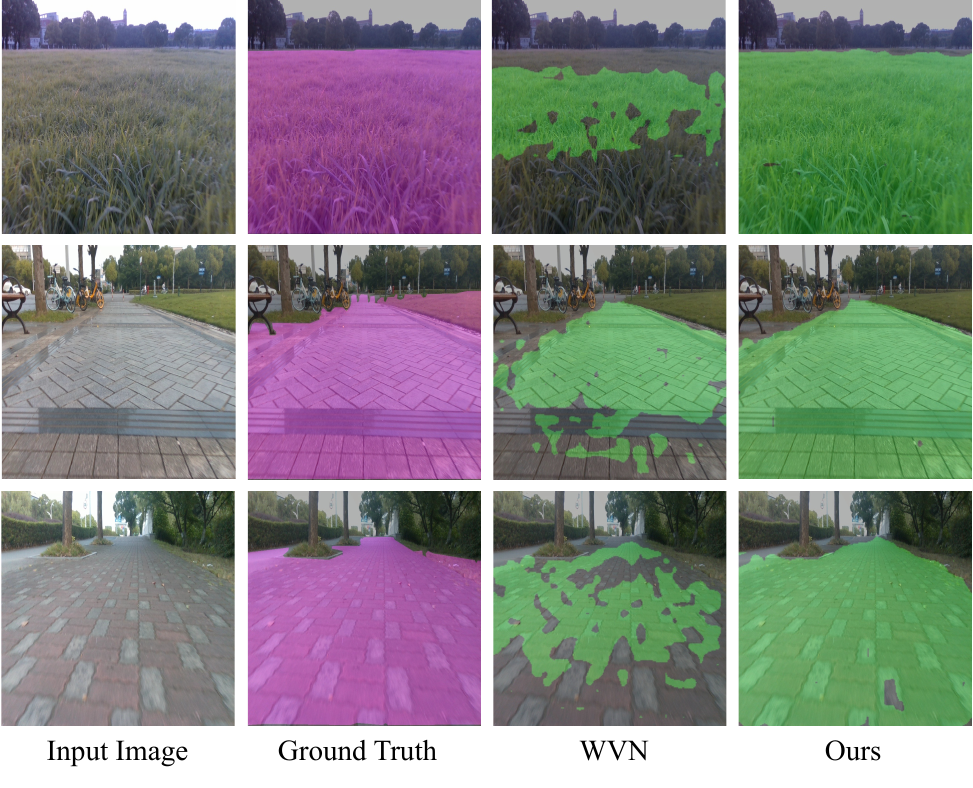}
    \caption{Qualitative results for self-collected dataset.}
    \label{fig:study3_demo}
\end{figure}

\subsubsection{Study 2} Evaluation on Public Dataset. Table \ref{table:study2_metric} and Fig.\ref{fig:study2_demo} present the quantitative and qualitative results for three short sequences. For testing the WVN algorithm, we used the weights provided by \cite{freyFastTraversabilityEstimation2023} for each scene. And the IMOST is trained online 300 steps. Experimental data indicates that both WVN and IMOST can quickly adapt to single scene. By analyzing the ROC curve, we find the optimal threshold $\alpha$ for each sequence. IMOST slightly outperforms WVN in terms of the AUROC metric, while WVN’s optimal threshold poses significant risk. Hence, F1 and IoU are calculated with a decision threshold of 0.5. Results demonstrate that IMOST outperforms WVN across all three sequences. Additionally, IDM generates only a small amount of memory data, thus efficiently conserving storage space.

In Fig.\ref{fig:study2_demo}, the purple area denotes the manually annotated traversable regions, while the green area represents the  algorithm’s predicted traversable areas. Our predictions exhibit superior completeness and accuracy in boundary distinction compared to WVN. The adaptation of WVN is notably affected by varying lighting conditions and different semantic scenes, whereas our method maintains high performance across these variations.

\begin{figure}[t]
    \centering
    \includegraphics[width=0.45\textwidth]{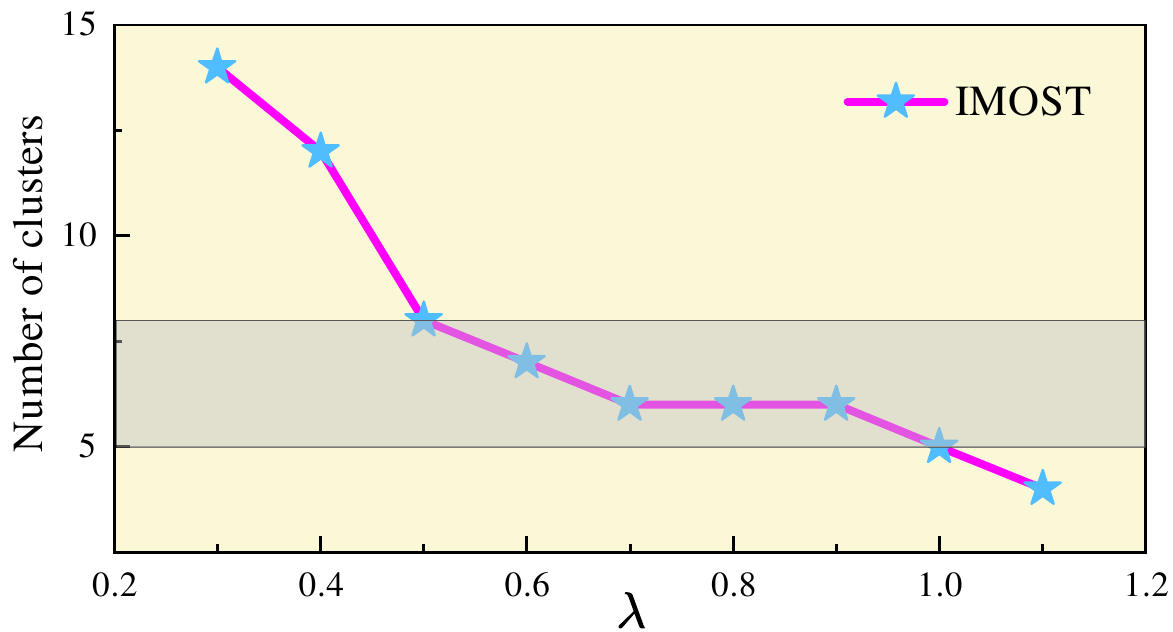}
    \caption{The number of memory clusters when changing $\lambda$.}
    \label{fig:study4_demo}
\end{figure}

\subsubsection{Study 3} Evaluation on Self-Collected Dataset. This sequence includes five types of terrain, as shown in Fig.\ref{fig:datastream}. Despite having no storage constraints, WVN’s performance drops significantly across all metrics. This decline is due to the distribution shift in semantic features, leading to catastrophic forgetting of previous scenes. During online learning, IDM creates 5 clusters and stores 100 images, while maintaining data balance between clusters according to capacity limitations. These aspects enable our method to continually learn traversability within complex scenarios. Qualitative results comparing the two methods are shown in Fig.\ref{fig:study3_demo}. The predictions of WVN are fragmented, and it does not achieve good distinction at the traversability boundaries. And recall rate of wvn is lower compared to our method. Our method accurately segments the traversable areas and aligns closely with the ground truth at the boundaries.

\subsubsection{Study 4} Ablation Study. We validate the performance of IDM and self-supervised annotation with FastSAM. According to the results from Table \ref{table:study3_metric}, the best performance is achieved when both IDM and FastSAM are used together, demonstrating the effectiveness of these two module. This combination significantly improves the algorithm’s recall rate. Additionally, we tested various values for the parameter $\lambda$. Five clusters are the optimal choice, balancing classification performance and storage requirements. As shown in Fig. \ref{fig:study4_demo}, the optimal range lies between 0.5 and 1.0.

\section{CONCLUSIONS}
In this paper, we propose IMOST for continual learning of traversability, which extends self-supervised methods. Inspired by the human memory expansion mechanism, IDM implements scene-aware data storage through increasing the clusters, ensuring the diversity of samples and balance of clusters. And annotation method assisted with FastSAM provides detailed boundries of traversability, which simplifies the difficulty of network learning. Both two modules improve the performance of traversability estimation and gain better adaptation to the task of continual learning. In future work, we plan to enhance robots’ active learning awareness in an embodied manner to reduce the computational burden of online learning.


\bibliographystyle{ieeetr}
\bibliography{References}

\addtolength{\textheight}{-12cm}   







\end{document}